 \documentclass[final,5p,times,twocolumn]{elsarticle}

\usepackage{amssymb}
\usepackage{lipsum}
\usepackage[colorlinks=true,linkcolor=blue,unicode=true]{hyperref}
\usepackage{euscript}
\usepackage{supertabular}

\usepackage{amsthm,amssymb, amsmath}
\usepackage{textcomp}
\usepackage[noend]{algorithmic}
\usepackage[ruled]{algorithm}
\usepackage{sectsty}
\usepackage{multirow}

\usepackage{diagbox}
\usepackage{indentfirst}

\journal{arXiv}

\begin{document}

\begin{frontmatter}



\title{Refining the ONCE Benchmark with Hyperparameter Tuning}


\author{\uppercase{Maksim Golyadkin}$^{a, b}$,
\uppercase{Alexander Gambashidze}$^{a, b}$,
\uppercase{Ildar Nurgaliev}$^{a}$,
\uppercase{Ilya Makarov}$^{a, b}$}

\address[1]{Artificial Intelligence Research Institute (AIRI), Moscow, Russia}
\address[2]{HSE University, Moscow, Russia}
\address[3]{AI Center, NUST MISiS, Moscow, Russia}

\begin{abstract}
In response to the growing demand for 3D object detection in applications such as autonomous driving, robotics, and augmented reality, this work focuses on the evaluation of semi-supervised learning approaches for point cloud data. The point cloud representation provides reliable and consistent observations regardless of lighting conditions, thanks to advances in LiDAR sensors. Data annotation is of paramount importance in the context of LiDAR applications, and automating 3D data annotation with semi-supervised methods is a pivotal challenge that promises to reduce the associated workload and facilitate the emergence of cost-effective LiDAR solutions. Nevertheless, the task of semi-supervised learning in the context of unordered point cloud data remains formidable due to the inherent sparsity and incomplete shapes that hinder the generation of accurate pseudo-labels. In this study, we consider these challenges by posing the question: "To what extent does unlabelled data contribute to the enhancement of model performance?"  We show that improvements from previous semi-supervised methods may not be as profound as previously thought. Our results suggest that simple grid search hyperparameter tuning applied to a supervised model can lead to state-of-the-art performance on the ONCE dataset, while the contribution of unlabelled data appears to be comparatively less exceptional.
\end{abstract}



\end{frontmatter}

%\tableofcontents


\section{Introduction}
\label{sec:introduction}

In an era where the interaction with 3D environments is increasingly necessary, there's a corresponding demand for refined deep learning technologies that cater to 3D data detection. Key domains such as autonomous driving, robotics, and augmented reality are just a few applications of this emerging need. Amidst various representations, our research focuses on the point cloud representation, primarily due to the unique benefits of the range sensor (LiDAR). Known for its consistent observations, LiDAR is unaffected by external parameters such as lighting conditions or the time of day, and exhibits resilience under various weather conditions \cite{solid_lidar}. In light of the anticipated rise of wearable devices equipped with compact LiDAR scanners \cite{wearable_sensors}, industry specialists anticipate a growing dependence on LiDAR \cite{lidar_mapping}.

The pressing challenge on the horizon relates to the demand for 3D data annotation, highlighting the increased importance of automated self-labeling data platform solutions within current and future LiDAR applications. Our research addresses this challenge by focusing on semi-supervised learning as a strategic approach to significantly reduce the 3D annotation workload (see \cite{ssl_survey} for more details). However, it is imperative to acknowledge that the field of semi-supervised learning presents its own formidable challenges, the most prominent of which are the creation of datasets and the establishment of validation methodologies.

The fundamental premise of semi-supervised approaches assumes that the labeled and unlabelled data have identical distributions. Nevertheless, the practical manifestation of this assumption often deviates from the ideal scenarios due to a lack of labeled data, which is insufficient to represent the entirety of the general population but is instead skewed towards specific modes. In the context of autonomous driving data, factors such as prevailing weather conditions or specific geographical locations may be disproportionately limited to a particular option. Hence, the utilization of datasets that can adequately account for these different characteristics becomes mandatory when validating semi-supervised learning methods.

We argue that the ONCE dataset, introduced recently, is the most appropriate option for fulfilling this requirement, serving as the prime benchmark for the comparative evaluation of semi-supervised 3D object detection models. Specifically, the current state-of-the-art approach, Proficient Teacher, relies predominantly on the ONCE dataset\cite{once} to support its claims of superiority over alternative methods. Nonetheless, it is worth mentioning that the authors of this study utilize training configurations and benchmarking results of previous methods provided within the ONCE data toolkit. Our empirical analysis reveals that these training parameters deviate significantly from optimality and that a properly tuned model trained exclusively on labeled data significantly outperforms Proficient Teacher \cite{ProfTeachers}.

The reason for this gap is that most semi-supervised learning methods rely on pseudo-labeling. A supervised model must be pretrained on labeled data to acquire initial pseudo labels. As a result, the effectiveness of semi-supervised learning depends on the quality of the pretrained model. Note that the models pretrained according to the ONCE benchmark tend to underfit, compromising the legitimacy of comparisons with semi-supervised methods. We argue that a fair assessment of semi-supervised techniques, particularly their effectiveness in learning from unlabelled data, can only be achieved with an adequately fitted pretrained model. Without this essential criterion, during semi-supervised training models would only undergo further training on labeled data, and the approach that hinders such training less would be superior to one that relies more heavily on unlabelled data.

Within this study, we aim to address the question: "How can we optimize the supervised pretraining process to provide a fairer comparison of semi-supervised methods?". Our research results in proper training hyperparameters for the SECOND \cite{second} and CenterPoint \cite{centerpoint} models, culminating in a significant improvement in the quality of supervised pretraining and, consequently, semi-supervised training. We present empirical evidence indicating that the difference between supervised and semi-supervised models is less significant than previously believed, implying sufficient scope for improvement within the domain of semi-supervised methods.

In summary, our contributions to this study encompass the following key aspects:
\begin{enumerate}
\item Enhancement of the SECOND Model: We have meticulously identified the optimal training parameters for the SECOND model, spanning the realms of pretraining, semi-supervised learning (SSL) training, and post-processing. This rigorous optimization process has led to a notable refinement of the benchmark results associated with this model.
\item Advancements in the CenterPoints Model: In the case of the CentrePoints model, our contributions extend to substantial improvements in the outcomes of the pretraining phase. Additionally, we have calculated metrics for semi-supervised methods, specifically Mean Teacher \cite{MeanTeachers} and Proficient Teacher \cite{ProfTeachers}, that had yet to be published.

\end{enumerate}

\section{Related works}
\subsection{Two-dimensional object detection}

In recent years, significant strides have been made in the field of object detection, a crucial component of computer vision. This advancement is largely credited to the development and enhancement of deep learning models. Despite common challenges posed by computer vision algorithms, such as object scale variation, occlusion, changes in lighting conditions, and the presence of previously unseen object categories, these algorithms are approaching human-like performance in certain scenarios. These algorithms primarily fall into two categories based on their architecture design: single-stage detectors and two-stage detectors. A series of publications \cite{rcnn, fast_rcnn, faster_rcnn} introduced the key principles of the two-stage detection framework \cite{rfcn, fpn, cascade_rcnn, libra_rcnn, htc, detectors}, which gradually replaced traditional methods with neural networks. Initially, a region proposal network is used to detect regions of interest (RoIs) that are likely to contain an object. These proposals are then refined in the second stage by predicting positional residuals for detected RoIs. Conversely, one-stage detectors (YOLO \cite{yolov1, yolov2, yolov3}, RetinaNet \cite{focal_loss}, SSD \cite{ssd}) directly predict bounding boxes and categories without refinement, offering faster execution but lower detection quality.

Further divisions in these categories include anchor-based and anchor-free methods. Anchor-based detectors like Faster R-CNN \cite{faster_rcnn} or YOLO \cite{yolov1} predict positional offset and scaling for anchors, pre-defined bounding boxes associated with different regions of the image. Anchor-free detectors \cite{densebox, fsaf, detr}, on the other hand, remove the need for pre-set anchor boxes and directly predict object boundaries from features such as keypoints \cite{reppoints}, center points \cite{fcos, fovea}, or extreme points \cite{cornernet, extremenet}. Despite anchor-based methods' early success and extensive literature base, anchor-free methods offer straightforwardness, flexibility, and eliminate the need for tuning anchor-related hyperparameters. Thus, these are gaining interest in the research community.

\subsection{Three-dimensional object detection}

The way 3D object detection methods categorize and predict bounding boxes mirrors that of the more mature 2D object detection field. However, the main differences lie in the backbone architectures and their extraction of information from sparse 3D data. These approaches can be broadly classified as voxel-based, point-based, and hybrid, which is a combination of the two.

Voxel-based methods \cite{pixor, pdv, sst, centernet, centerformer, liga_stereo, p2s, vista, btc} convert point clouds into regular voxel grids that are processed using 3D convolutions. This method introduces a structure absent in point clouds and simplifies processing. It also allows for the application of 2D convolutional neural network architectures. Nonetheless, the direct application of voxelization and 3D convolutions as proposed in VoxelNet \cite{voxelnet} results in considerable computational demands due to cubic complexity from voxelization. This also results in the loss of fine-grained details and quantization artifacts. Furthermore, most voxel grids generated from typical point clouds are empty, making these computations largely unnecessary. To address this, SECOND \cite{second} proposed sparse convolutions to avoid processing empty voxels. PointPillars \cite{pointpillars} simplifies voxel representations to pillar representations, or BEV maps, to maintain high efficiency and competitive performance. SA-SSD introduces an auxiliary network to exploit point-wise supervision.

Point-based methods \cite{lidar_rcnn, 3dssd, ia_ssd} operate directly on raw point clouds to extract pointwise features. The model PointNet \cite{pointnet} introduced this framework, followed by its successor PointNet++ \cite{pointnetpp}, to process points directly without losing structural information due to data representation. Frustum-PointNet \cite{frustum_pointnet} uses 2D detection to constrain space with the frustum corresponding to the 2D bounding box. PointRCNN \cite{point_rcnn} ideologically follows Faster-RCNN but generates proposals directly from points. VoteNet \cite{votenet} uses voting to predict proposal centers from point clusters and mitigate the occlusion problem. These methods, however, can experience computational inefficiencies and limited learning capacity due to their reliance on operations such as kNN search or k-d tree construction because raw point clouds lack the structural information needed for neural networks to perform computations.

Hybrid methods \cite{vst, fpt} strive to combine the strengths of voxel-based and point-based approaches while ensuring efficiency. STD \cite{std} employs PointNet++ to identify the most relevant points in the sparse representation and converts them to a dense representation processed with convolutions. PV-RCNN \cite{pv_rcnn} refines proposals using pointwise features. HVPR \cite{hvpr} incorporates both voxel-wise and pointwise features into a single 3D representation with a memory module.

\subsection{Semi-supervised learning}

Semi-supervised learning leverages the potential of unlabeled data to enhance classification accuracy. Since images offer ample meaningful information about the underlying data distribution even without labels, these primary approaches in semi-supervised learning – self-training and consistency regularization – with multiple algorithms developed within each approach, both smooth the feature manifold \cite{smoothness} by stabilizing predictions and increasing confidence for unlabeled data. This process assists models in learning robust decision boundaries, improving performance, generalization, and overall quality without the need for costly manual annotation.

Self-training \cite{MeanTeachers, noisy_student, fixmatch, conf_bias, simmatch, cossl} begins with initial training on labeled data, followed by iterative generation of high-quality pseudo-labels for unlabeled data and training on the expanded labeled dataset. Improving the performance of these methods is associated with maximizing valuable information obtained from pseudo-labels and minimizing the mislabeling produced by the overfitted model. Consistency regularization \cite{temp_ensembling, mixmatch} maintains consistent predictions for the input under various perturbations, mainly implemented with image augmentations. The research on these methods is related to finding the optimal loss that provides consistency and ways to acquire perturbations.

\subsection{Semi-supervised learning for object detection}

Applying semi-supervised methods to more complex tasks such as object detection presents challenges due to the complex label assignment associated with the multi-task predictions of object detectors. While a few consistency-based methods \cite{csd, isd} enforcing consistency between augmented views have been proposed for semi-supervised 2D object detection, self-training approaches are currently more prevalent. The teacher-student framework-based approach was initially proposed in STAC \cite{stac}, where the teacher, pretrained on labeled data, predicts pseudo-labels for weakly augmented views of unlabeled data, and the student learns to predict

%%%%%%%

\section{Models and Hyperparameters}
In this section, we briefly describe SECOND and CenterPoint models used for 3D object detection and describe which training hyperparameters we choose to tune and why they are essential. 
{\centering{The overview of the SECOND \cite{second} detector.The detection process commences with the utilization of an initial raw point cloud as the input data. Subsequently, this point cloud undergoes a  conversion into voxel-based features and coordinates. The transformed data then traverses through two layers of VFE (voxel feature encoding) followed by a linear layer. A sparse convolutional neural network (CNN) is subsequently applied to further process the encoded features. Finally, the detection process is completed with the operation of a Region Proposal Network (RPN) responsible for generating the final detections.
\label{fig:second}}}

We follow the \cite{ioumatch} and \cite{ProfTeachers}, the most prominent works in semi-supervised 3D object detection, and elaborate experiments with the SECOND and CenterPoint detectors. These models may not be the latest or most sophisticated, yet their simplicity permits the isolation and highlighting of the effects of semi-supervised learning. Additionally, they support real-time operation, a critical requirement for the majority of applications.

\textbf{SECOND} (Sparsely Embedded Convolutional Detection) is primarily designed for autonomous driving and robotics use. Its architecture consists of the fusion of three core components: a Voxel Feature Extractor, a sparse convolutional middle layer, and a region proposal network. The Voxel Feature Extractor (VFE) transforms raw point cloud data into structured representation by dividing it into fixed-size 3D grids called "voxels." SECOND extracts meaningful features containing spatial and semantic information within each voxel. These features are then combined to produce a feature map. Sparse 3D convolutions, which are significantly faster and require less memory than their original counterparts, are used to process the feature map. Finally, RPN uses the feature map generated by the convolutional layers to produce 3D object proposals by predicting their 3D bounding boxes and associated objectness scores. After the RPN proposes objects, an essential post-processing step is taken to refine the detected objects. Non-maximum suppression (NMS) is frequently used to eliminate redundant or significantly overlapping proposals, leaving only the most confident nonintersecting objects.

\textbf{CenterPoint}. The core of this architecture (see Fig. \ref{fig:center}) is the Center heatmap head. It recognizes probable object centers within the point cloud for each class independently. Once centers are identified, CentrePoint predicts the 3D bounding boxes for each object by regressing its dimensions and orientations concerning the centers. Subsequently, in the second stage, boxes are refined by processing features interpolated for center position with MLP. This approach simplifies the process, making it computationally efficient. Similar to SECOND, post-processing includes NMS and score thresholding.

\begin{table*}[!ht]
\caption{Grid search results on validation for SECOND trained on labeled data split. For every cell, pair corresponds to mAP values for the original NMS threshold and the tuned one. Metrics demonstrate better results for higher values (best metrics are highlighted in bold).}
\label{gs_tab_second}
\centering{
\begin{tabular}{|c|c|c|c|c|c|c|}
\hline
\diagbox{Batch size}{Learning rate} & 0.0001 & 0.0005 & 0.001   & 0.003 & 0.006  & 0.01           \\
\hline
\multicolumn{7}{c}{80 epochs}        \\
\hline
16                 & 50.60 / 51.56 & 51.17 / 51.78 & 52.32 / 53.67   & 52.11 / 53.51 & 51.76 / 52.13  & 49.87 / 50.49   \\
32                  & 50.11 / 50.92 & 50.76 / 51.42 & 51.87 / 53.20   & 51.95 / 53.14 & 51.74 / 52.78  & 50.11 / 51.37   \\
64                  & 49.55 / 50.24 & 51.23 / 52.03 & 52.96 / 53.78   & 53.47 / 54.29 & 53.55 / 54.41  & 50.99 / 51.78   \\
128                 & 49.12 / 49.89 & 49.82 / 50.76 & 51.81 / 52.43   & 53.89 / 54.48 & 54.08 / 55.23  & 52.17 / 52.84   \\
\hline
\multicolumn{7}{c}{150 epochs}        \\
\hline
16                 & 50.45 / 51.32 & 52.02 / 52.63 & 52.77 / 53.92   & 52.65 / 53.82 & 52.27 / 53.02  & 49.76 / 50.38   \\
32                  & 51.57 / 52.34 & 52.37 / 54.01 & 54.25 / 55.54   & 55.17 / 57.24 & 55.11 / 57.13  & 52.17 / 53.34    \\
64                  & 51.98 / 52.76 & 52.68 / 54.31 & 54.92 / 56.11   & 55.46 / 57.41 & 56.28 / 57.82  & 52.87 / 54.02   \\
128                 & 52.34 / 53.12 & 52.89 / 54.02 & 54.87 / 55.91   & 55.78 / 57.81 & 56.65 / 58.09  & 53.21 / 54.42   \\
\hline
\multicolumn{7}{c}{500 epochs}        \\
\hline
16                 & 51.32 / 52.18 & 57.15 / 59.69 & 57.97 / 59.26   & 58.12 / 60.76 & 54.05 / 55.48 & 51.98 / 53.04   \\
32                  & 52.06 / 53.22 & 56.78 / 58.41 & 57.92 / 59.21   & 58.47 / 60.03 & 56.39 / 57.87 & 53.21 / 54.36   \\
64                  & 53.47 / 54.98 & 58.75 / 59.11 & 59.07 / 60.75   & 60.57 / 61.43 & 60.43 / 61.78  & 52.10 / 53.13   \\
128                  & 53.92 / 54.79 & 57.48 / 58.86 & 59.98 / 61.21   & 60.73 / 62.38 & 60.87 / 62.43 & 54.12 / 55.38   \\
\hline
\multicolumn{7}{c}{1000 epochs}        \\
\hline
16                 & 51.76 / 52.64 & 57.43 / 59.12 & 58.88 / 60.17   & 58.12 / 59.08 & 59.07 / 59.48 & 53.09 / 54.23   \\
32                  & 52.18 / 53.32 & 56.97 / 58.61 & 58.14 / 59.43   & 59.19 / 61.25 & 60.61 / 60.05 & 53.73 / 54.88   \\
64                  & 53.14 / 54.29 & 58.78 / 59.32 & 58.67 / 59.95   & 60.28 / 61.86 & 60.21 / 61.73 & 54.36 / 55.51   \\
128                  & 51.15 / 51.99 & 58.12 / 59.26 & 58.44 / 59.30   & 61.45 / 62.96 & 61.97 / \textbf{63.22} & 55.14 / 56.23   \\
\hline
\end{tabular}
}
\end{table*}

\begin{table*}[!ht]
\caption{Grid search results on validation for CenterPoint trained on labeled data split. For every cell, pair corresponds to mAP values for the original NMS threshold and the tuned one. Metrics demonstrate better results for higher values (best metrics are highlighted in bold).}
\label{gs_tab_center}
\centering{
\begin{tabular}{|c|c|c|c|c|c|c|}
\hline
\diagbox{Batch size}{Learning rate} & 0.0001 & 0.0005 & 0.001   & 0.003 & 0.006  & 0.01           \\
\hline
\multicolumn{7}{c}{80 epochs}        \\
\hline
16                 & 52.30 / 52.87 & 52.49 / 52.82 & 54.47 / 54.88   & 53.08 / 53.42 & 52.87 / 53.29  & 50.12 / 50.50   \\
32                  & 51.92 / 52.41 & 53.08 / 53.51 & 54.83 / 55.39   & 54.95 / 55.49 & 54.83 / 55.22  & 50.91 / 51.27   \\
64                  & 50.85 / 51.43 & 53.19 / 53.43 & 55.05 / 55.63   & 55.72 / 55.13 & 55.81 / 56.37  & 51.20  / 51.62   \\
128                 & 50.32 / 50.74 & 53.15 / 53.24 & 55.61 / 56.07   & 55.93 / 56.33 & 55.78 / 56.15  & 51.57 / 51.93   \\
\hline
\multicolumn{7}{c}{150 epochs}        \\
\hline
16                 & 52.15 / 52.64 & 52.43 / 52.98 & 56.62 / 57.09   & 57.25 / 57.78 & 56.98 / 57.52  & 50.01 / 50.39   \\
32                  & 53.27 / 53.87 & 54.47 / 55.01 & 57.26 / 57.83   & 58.11 / 58.56 & 57.29 / 57.63  & 53.07 / 53.41    \\
64                  & 52.89 / 53.46 & 54.59 / 55.12 & 57.87 / 58.31   & 58.32 / 58.78 & 58.21 / 58.63  & 52.86 / 53.22   \\
128                 & 52.78 / 53.33 & 54.68 / 55.23 & 58.03 / 58.47   & 58.49 / 58.94 & 59.39 / 59.81  & 53.15 / 53.51   \\
\hline
\multicolumn{7}{c}{500 epochs}        \\
\hline
16                 & 52.54 / 53.01 & 57.02 / 57.57 & 60.47 / 60.92  & 60.52 / 60.97 & 58.45 / 58.88 & 52.38 / 52.73   \\
32                  & 53.10 / 53.57 & 58.58 / 59.13 & 59.98 / 60.43   & 61.43 / 61.87 & 60.25 / 60.78 & 53.22 / 53.57   \\
64                  & 52.82 / 53.17 & 59.24 / 59.67 & 60.41 / 61.09   & 62.18 / 62.67 & 62.74 / 63.26  & 53.02 / 53.38   \\
128                  & 53.37 / 53.82 & 59.78 / 60.23 & 61.08 / 61.53   & 62.53 / 62.98 & 62.43 / 62.88  & 54.19 / 54.54   \\
\hline
\multicolumn{7}{c}{1000 epochs}        \\
\hline
16                 & 52.76 / 53.21 & 58.23 / 58.78 & 60.61 / 61.07   & 60.07 / 60.53 & 59.01 / 59.46 & 50.94 / 51.29   \\
32                  & 53.14 / 53.59 & 59.64 / 60.09 & 60.97 / 61.43   & 61.42 / 61.87 & 60.34 / 60.78 & 51.38 / 51.73   \\
64                  & 53.14 / 53.69 & 59.78 / 60.12 & 60.67 / 61.35   & 62.28 / 62.86 & 61.21 / 61.73 & 52.36 / 52.51   \\
128                  & 52.53 / 52.91 & 60.87 /61.26 & 61.28 / 62.17   & 63.88 / \textbf{64.41} & 63.57 / 64.12 & 54.14 / 55.07   \\
\hline
\end{tabular}
}
\end{table*}

We perform a straightforward grid search operation to maximize the quality of training and inference for these models. We investigate the influence of the following components:

\begin{itemize}
    \item \textbf{Learning rate} is typically one of the first hyperparameters to be tuned. It sets the size of the steps taken during the optimization process when updating the model's parameters based on the gradient of the loss function. It influences the neural network's speed of convergence for solution, regulates numerical stability, and affects the model's generalizability to new, unseen data.
    \item \textbf{Batch size} determines the number of training examples used in each iteration of the optimization process, impacting both the training dynamics and the computational requirements. Smaller batch sizes introduce more randomness into the optimization process, allowing the model to escape local minima and explore a wider range of solutions. However, using very small batch sizes may also lead to noisy updates and hinder convergence. Larger batch sizes can result in more stable updates but may also lead to overfitting if regularization is not provided. Additionally, batch size and learning rate jointly affect training \cite{lr_bs}. Smaller batch sizes typically necessitate a lower learning rate because each batch provides less information about the overall dataset, and the model's weights must be updated more carefully to avoid instability. Conversely, a larger batch size can accommodate a larger learning rate as it offers a more stable estimate of the gradient.
    \item \textbf{Number of epoch} defines how many times the entire training dataset is processed. Choosing the right number of epochs balances model convergence and prevents overfitting, determined by monitoring validation performance. It also influences computational resources and should be considered alongside other hyperparameters like learning rate and batch size.
    \item \textbf{NMS threshold} is a post-processing parameter that determines the merging or suppression of overlapping or closely spaced 3D bounding boxes. It specifies the maximum allowed overlap between two detections required to be considered separate and corresponds to the 3D IoU metric. The selection of non-maximum suppression (NMS) threshold significantly affects the precision and recall of the object detection system. Raising the threshold boosts precision but may compromise recall by discarding valid detections, whereas lowering it might enhance recall but could lead to more false positives.
\end{itemize}

{\centering{The overview of the CenterPoint \cite{centerpoint} detector. Initially, a conventional 3D backbone extracts map-view feature representations from lidar point cloud data. Subsequently, a specialized 2D convolutional neural network integrated into the detection head identifies object centers and performs a regression to determine the complete 3D bounding boxes based on these center features. The predicted bounding box information is then used to locate and extract point-based features at the 3D centers of each face of the estimated 3D bounding box. These extracted features are then fed into a Multilayer Perceptron to project a confidence score according to IoU and refine the bounding box regression.\label{fig:center}}}

\section{Results}

This section presents a comprehensive analysis of various semi-supervised learning methods using the ONCE dataset, demonstrating that our hyperparameter setup achieves state-of-the-art performance among existing semi-supervised methods for SECOND and CenterPoint models. Firstly we provide a detailed description of the ONCE dataset and highlight its unique features in Section \ref{dataset}. Secondly, we outline our experiments' training methodology and configurations in Section \ref{training_setup}. Finally, we compare the performance of different semi-supervised methods using the mean Average Precision (mAP) metric in Section \ref{comparison}.

\subsection{ONCE dataset}
\label{dataset}

The ONCE \cite{once} dataset is a large-scale autonomous driving dataset consisting of 1 million LiDAR point cloud samples corresponding to 144 hours of driving across different cities in China. The ONCE dataset stands out among open-source autonomous driving datasets due to its larger size and greater diversity in weather and traffic conditions. This diversity is essential for training and evaluating autonomous driving systems, as it enables the development of more robust and versatile models. The detection task focuses on five foreground object classes: Car, Bus, Truck, Pedestrian, and Cyclist. However, during evaluation, the classes Car, Bus, and Truck are combined into a single "Vehicle" class.

The ONCE dataset is designed to evaluate semi-supervised and self-supervised learning approaches for 3D object detection. It contains 581 sequences with 20 labeled and 561 unlabeled sequences. The labeled data is divided into training, validation, and test splits. The splits are organized as follows:
\begin{itemize}
    \item Training split: 6 sequences (5k scenes) captured on sunny days
    \item Validation split: 4 sequences (3k scenes) with diverse weather conditions - 1 sunny day, 1 rainy day, 1 sunny night, and 1 rainy night
    \item Testing split: 10 sequences (8k scenes) covering various weather conditions - 3 sunny days, 3 rainy days, 2 sunny nights, and 2 rainy nights
\end{itemize}

Both downtown and suburban areas are covered in each split. The training split has a slight domain shift compared to the validation/testing split to encourage better generalizability of the proposed methods.

The remaining 560 sequences are kept as unlabeled data for research on leveraging large-scale unlabeled data. These unlabeled scenes are divided into three subsets: Small, Medium, and Large.
\begin{itemize}
    \item Small: 70 sequences (100k scenes)
    \item Medium: 321 sequences (500k scenes)
    \item Large: 560 sequences (about 1M scenes)
\end{itemize}

Small is a subset of Medium, which is a subset of Large. Small and Medium are created by selecting specific roads in time order rather than uniformly sampling from all scenes.

The performance of 3D object detectors is evaluated using mean Average Precision (mAP) \cite{pascalvoc} over all classes, based on 3D Intersection over Union (IoU) thresholds of 0.7, 0.3, and 0.5 for Vehicle, Pedestrian, and Cyclist classes, respectively. Additionally, the detector's performance is assessed over three different detector ranges: 0-30m, 30-50m, and 50m-inf.

\begin{table*}[!ht]

\caption{Evaluation results on ONCE validation split for SECOND trained on different splits of unlabeled data (Small, Medium, Large). Models marked with * are initialized with SECOND*, while others are initialized with SECOND. Metrics demonstrate better results for higher values (best metrics are highlighted in bold, and second best are underlined). Our hyperparameter setup exhibits state-of-the-art performance in terms of mAP for every split.}
\label{results_tab_second}

\resizebox{.95\textwidth}{!}{
\begin{tabular}{|l|cccc|cccc|cccc|c|}
\hline
\multirow{2}{*}{Method} & \multicolumn{4}{c|}{Vehicle AP}     & \multicolumn{4}{c|}{Pedastrian AP}  & \multicolumn{4}{c|}{Cyclist AP}     & \multirow{2}{*}{mAP} \\
                        & overall & 0-30m & 30-50m & 50m-inf & overall & 0-30m & 30-50m & 50m-inf & overall & 0-30m & 30-50m & 50m-inf &                      \\
\hline
\multicolumn{14}{c}{\textbf{Training} (5k labeled samples)}        \\
\hline
SECOND                  & 71.19   & 84.04 & 63.02  & 47.25   & 26.44   & 29.33 & 24.05  & 18.05   & 58.04   & 69.96 & 52.43  & 34.61   & 51.89                \\
SECOND*                &    78.22     &  87.57     &   73.43     &    59.16     &   43.92 &   52.56     &    35.65      &   19.61      &   67.52    &    77.92    &    63.18     &         45.12  &     63.22      \\
\hline
\multicolumn{14}{c}{\textbf{Small} (5k labeled + 100k unlabeled samples)}        \\
\hline

Pseudo Label            &  72.80 &     84.46  &    64.97    &     51.46    &    25.50     &    28.36   &    22.66    &    18.51     &    55.37     &  65.95     &     50.34    &     34.42    &            51.22          \\
Noisy Student           &   73.69      &   84.69    &     67.72    &      53.41    &   28.81      &    33.23    &   23.42      &    16.93     &     54.67    &     65.58   &      50.43   &     32.65    &       52.39               \\
Mean Teacher            &   74.46      &  86.65     &   68.44     &     53.59    &    30.54     &   34.24    &  26.31      &    20.12      &    61.02     &  72.51     &   55.24     &   39.11      &        55.34              \\
SESS                    &  73.33       &   84.52     &  66.22      &   52.83      &    27.31     &    31.11    &    23.94     &    19.01     & 59.52         &  71.03    &   53.93      &    36.68     &          53.39           \\
3DIoUMatch              &     73.81    &   84.61     &    68.11    &     54.48     &   30.86      & 35.87      & 25.55       &   18.30      &      56.77   &  68.02     &   51.80     &  35.91       &      53.81                \\
Proficient Teacher      &  76.07       &  \underline{86.78}    &   70.19     &  56.17       &   35.90      & 39.98      &  31.67      &    \underline{24.37}       &  61.19      &   73.97    &   55.13    &    36.98       &        57.72            \\
Mean Teacher*          &   \underline{77.61}      &   86.70    &  \underline{73.31}       &  \underline{59.37}      &    \underline{46.40}     &   \underline{53.36}    &   \underline{41.24}     &    23.19     &    \underline{68.92}     &   \underline{79.37}    &    \underline{63.65}    &   \underline{47.55}      &       \underline{64.31}                \\
Proficient Teacher*    &    \textbf{80.59}     &  \textbf{89.02}    &  \textbf{76.64}      &  \textbf{63.33}       &   \textbf{50.58}      &  \textbf{59.27}     &    \textbf{43.13}    &   \textbf{24.99}      &  \textbf{69.75}       &  \textbf{80.80}      &   \textbf{65.03}     &    \textbf{47.57}      &        \textbf{67.06}               \\

\hline
\multicolumn{14}{c}{\textbf{Medium} (5k labeled + 500k unlabeled samples)}      \\
\hline
Pseudo Label            & 73.03  &     86.06   &   65.96     &  51.42        &   24.56      &  27.28     &     20.81   &  17.00       &    53.61       &   65.26   &   48.44    &   33.58      &           50.40           \\
Noisy Student           &   75.53      &   86.52    &   69.78      &    55.05      &   31.56      &  35.80     &   26.24     &   21.21      &  58.93       &  69.61     &    53.73    &     36.94    &      55.34                \\
Mean Teacher            &   76.01      &   86.47    &  70.34      &    55.92     &   35.58      &    40.86   &   30.44     &   19.82      &    63.21     &   74.89    &   56.77     &     40.29     &      58.27                \\
SESS                    &  72.11       &   84.06    &   66.44     &  53.61       &   33.44      &  38.58      &   28.10    &    18.67     &    61.82     &    73.20   &    56.60    &     38.73     &         55.79             \\
3DIoUMatch              &  75.69       &  86.46     &   70.22      &  56.06       &  34.14       &  38.84     &   29.19      &    19.62      &     58.93    &  69.08     &    54.16    &    38.87      &       56.25               \\
Proficient Teacher      &   78.07      &   87.43    &    72.50    &   59.51      &   38.38      &   42.45    &    34.62    &  25.58       &   63.23      &  74.70     &    58.19    &    40.73     &    59.89                  \\
Mean Teacher*         &   \underline{78.81}      &  \underline{88.21}    &  \underline{74.26}      & \underline{60.03}       &   \underline{46.45}     &   \underline{54.17}     &  \underline{38.72}      &    \underline{22.68}     &   \underline{68.95}      &   \underline{79.73}   &    \underline{63.39}     &    \underline{46.56}     &        \underline{64.73}              \\
Proficient Teacher*     &    \textbf{80.08}     &  \textbf{88.30}      &  \textbf{76.43}      &  \textbf{63.85}       &    \textbf{52.54}     & \textbf{60.73}      &  \textbf{44.35}     &   \textbf{26.08}    &     \textbf{69.86}    &   \textbf{80.20}    & \textbf{64.98}      &   \textbf{48.22}      &        \textbf{67.49}             \\

\hline
\multicolumn{14}{c}{\textbf{Large} (5k labeled + 1M unlabeled samples)}      \\
\hline
Pseudo Label            &  72.41 &    84.06    &     64.54   &   50.05      &   23.62      &       26.80     &     20.13     &    16.66      &   53.25    &    64.69     &   48.52      &      33.47 &        49.76          \\
Noisy Student           &  75.99       & 86.67      &  70.48      &   55.60      &   33.31      &  37.81      &   28.19      & 21.39        &   59.81      &  70.01     &    55.13     &   38.33       &      56.37                \\
Mean Teacher            &    76.38     &   86.45     &    70.99     &   57.48      &   35.95      &   41.76    &   29.05     &    18.81      &      65.50   &  75.72      &   60.07      &   43.66       &         59.28             \\
SESS                    &  75.95      &   86.83    &    70.45     &  55.76       &  34.43       &    40.00    &     27.92    &     19.20     &      63.58   &    74.85   &    58.88     &    39.51      &       57.99               \\
3DIoUMatch              &    75.81     &   86.11     &  71.82       &    57.84      &    35.70     &   40.68    &    30.34    &   21.15      & 59.69        &    70.69   &     54.92    &   39.08      &         57.07             \\
Proficient Teacher      &    78.12     & 87.22     &  72.74      &  59.58       &  41.95       &    48.09   &  35.13      &   \underline{26.01}      &   64.12      &  75.85     &  58.04      &  41.45       &        61.40              \\
Mean Teacher*        &   \underline{79.07}      &    \underline{87.47}   &   \underline{74.24}     &     \underline{61.48}    &    \underline{46.98}     &   \underline{54.42}   &    \underline{39.11}      &  25.01       &   \underline{69.65}      &  \underline{79.85}    &   \underline{59.43}     &    \underline{42.35}     &        \underline{65.03}              \\
Proficient Teacher*   &    \textbf{81.13}    &   \textbf{88.55}    &   \textbf{75.75}     &   \textbf{64.81}     &     \textbf{52.87}    &  \textbf{61.15}     &    \textbf{45.23}    &  \textbf{26.63}       &      \textbf{70.85}   &  \textbf{80.97}    &   \textbf{65.76}    &   \textbf{49.13}      &           \textbf{67.89}           \\

\hline
\end{tabular}
}
\end{table*}

\begin{table*}[!ht]
\caption{Evaluation results on ONCE validation split for CenterPoint trained on different splits of unlabeled data (Small, Medium, Large). Models marked with * are initialized with CenterPoint*, while others are initialized with CenterPoint. Metrics demonstrate better results for higher values (best metrics are highlighted in bold, and second best are underlined). Our hyperparameter setup exhibits state-of-the-art performance in terms of mAP for every split.}
\label{results_tab_center}
\resizebox{.95\textwidth}{!}{
\begin{tabular}{|l|cccc|cccc|cccc|c|}
\hline
\multirow{2}{*}{Method} & \multicolumn{4}{c|}{Vehicle AP}     & \multicolumn{4}{c|}{Pedestrian AP}  & \multicolumn{4}{c|}{Cyclist AP}     & \multirow{2}{*}{mAP} \\
                        & overall & 0-30m & 30-50m & 50m-inf & overall & 0-30m & 30-50m & 50m-inf & overall & 0-30m & 30-50m & 50m-inf &                      \\
\hline
\multicolumn{14}{c}{\textbf{Training} (5k labeled samples)}        \\
\hline
CenterPoint                  & 66.79   & 80.10 & 59.55  & 43.39   & 49.90   & 56.24 & 42.61  & 26.27   & 63.45   & 74.28 & 57.94  & 41.48   & 60.05                \\
CenterPoint*                &    73.27     &  85.27     &   67.28     &    51.74     &   51.86           &   63.70     &    40.23      &   19.53      &   68.11    &    78.54    &    62.99     &        45.16  &    64.41      \\
\hline
\multicolumn{14}{c}{\textbf{Small} (5k labeled + 100k unlabeled samples)}        \\
\hline
Mean Teacher            &   67.12      & 79.95     &   60.23     &     42.73    &    52.84     &   58.13    & 43.28      &    26.95      &    66.01     &  75.99     &   58.86     &   42.33      &        62.34              \\
Proficient Teacher      &  72.09       &  84.68    &   66.19     &  50.11       &  54.30      & 64.52      &  41.36      &    \underline{27.08}       &    67.72    &   76.59    &   60.11    &    43.18       &        65.98            \\
Mean Teacher*          &   \underline{74.07}     &   \underline{85.00}    &  \underline{69.45}       &   \underline{54.87}      &    \underline{56.17}     &   \underline{66.31}    &  \underline{45.02}     &    26.51     &    \underline{69.33}     &   \underline{80.11}    &    \underline{63.78}    &   \underline{46.12}      &       \underline{66.47}                \\
Proficient Teacher*    &    \textbf{75.99}      &   \textbf{86.51}    &  \textbf{71.35}       &   \textbf{56.51}      &    \textbf{56.26}     &   \textbf{66.45}    &   \textbf{45.69}     &    \textbf{27.23}     &    \textbf{70.90}     &   \textbf{81.65}    &    \textbf{65.31}    &   \textbf{48.07}      &       \textbf{67.72}                \\
\hline
\multicolumn{14}{c}{\textbf{Medium} (5k labeled + 500k unlabeled samples)}      \\
\hline
Mean Teacher            &   68.20      & 80.45     &   61.34     &     43.95    &    53.12     &   58.67    & 44.02      &    27.38      &    66.75     &  76.31     &   59.33     &   42.89      &        63.80              \\
Proficient Teacher      &  \underline{76.11}       &  \textbf{88.14}    &   \underline{70.42}     &  53.95       &  \textbf{57.58}      & \textbf{67.83}      &  43.79      &    \textbf{30.31}       &    \underline{69.98}    &   78.91    &   62.57    &    45.88       &       \underline{68.22}            \\
Mean Teacher*          &   74.45     &   85.36    &  69.78       &   \underline{55.12}      &    56.45     &   66.76    &  \underline{45.37}     &    \underline{26.84}     &    69.88    &   \underline{80.42}    &    \underline{64.04}    &   \underline{46.89}      &       66.87              \\
Proficient Teacher*     &    \textbf{77.47}     &  \underline{86.88}      &  \textbf{72.19}      &  \textbf{58.20}       &    \underline{57.30}     & \underline{67.79}      &  \textbf{47.03}     &   25.66    &     \textbf{70.84}   &  \textbf{81.53}    & \underline{65.26}       &   \textbf{47.87}      &        \textbf{68.54}              \\
\hline
\multicolumn{14}{c}{\textbf{Large} (5k labeled + 1M unlabeled samples)}      \\
\hline
Mean Teacher            &    69.50     &   81.25     &    62.78     &  45.15      &  53.48     &  59.12   &  44.67     &    27.99      &    67.12   &  76.72    &  60.33    &  43.45      &         64.80             \\
Proficient Teacher      &    \underline{76.23}     &  \textbf{87.98}     &  \underline{70.64}     &   \underline{55.91}      &  \textbf{57.12}     &  66.43    &  43.63     &    \textbf{30.12}      &    \textbf{70.37}   &  \underline{80.51}    &  63.19    &  46.22      &         \underline{68.48}             \\
Mean Teacher*          &   75.78     &   85.64    &  69.93       &   55.42      &    56.78     &   \underline{67.12}    &  \underline{46.29}     &    27.82     &    69.81     &   79.91    &    \underline{63.92}    &   \underline{47.81}      &       67.45                \\
Proficient Teacher*    &    \textbf{78.49}     &   \underline{87.56}    &  \textbf{72.36}       &   \textbf{57.88}      &    \underline{57.02}     &   \textbf{67.49}    &   \textbf{47.25}     &    \underline{28.33}     &    \underline{70.23}     &   \textbf{80.74}    &    \textbf{64.05}    &   \textbf{48.36}      &       \textbf{69.68}                \\
\hline
\end{tabular}
}
\end{table*}

\subsection{Training setup}
\label{training_setup}

To ensure the reproducibility of our experiments, we use the codebase of the official ONCE benchmark \footnote{\url{https://github.com/PointsCoder/ONCE_Benchmark}}, which is based on OpenPCDet. The Proficient Teacher code was obtained from the authors and is also based on the ONCE benchmark. The experiments are performed on sixteen NVIDIA A100 80 GB GPUs.

First, we trained the SECOND and CenterPoints models from scratch using the hyperparameter configurations provided with the ONCE benchmark, which are the same for both models. Namely, training was performed on the training split for 80 epochs with a batch size of 32 and a maximum learning rate of 0.003 under the One Cycle learning rate policy \cite{one_cycle}. The NMS threshold during inference was also identical and was 0.01.

After that, we discovered that the proposed hyperparameters were suboptimal, and models were underfitted. Consequently, a grid search over batch size, learning rate, number of epochs, and NMS threshold was conducted to identify parameters that yielded significantly better results. The considered values and corresponding metric values are exhibited in Table \ref{gs_tab_second} and Table \ref{gs_tab_center} for SECOND and CenterPoint correspondingly. With this operation, we found out the set of optimal hyperparameters. It is batch size 128, learning rate 0.006, 1000 epochs, and 0.65 NMS threshold for the SECOND detector, and batch size 128, learning rate 0.003, 1000 epochs, and 0.25 NMS threshold for CenterPoint.
We used original hyperparameters of the ONCE benchmark for every semi-supervised learning approach.

\subsection{Model comparison}
\label{comparison}

The ONCE benchmark codebase provides the implementation of three image-based semi-supervised methods: Pseudo Label \cite{pseudolabel}, Mean Teacher \cite{MeanTeachers}, and Noisy Student \cite{noisy_student}, as well as two semi-supervised methods for point cloud detection: SESS \cite{sess} (designed for indoor datasets) and 3DIoUMatch \cite{ioumatch} (both indoor and outdoor). We perform semi-supervised learning for pretrained detectors (with original and tuned hyperparameters) using Mean Teacher and Proficient Teacher approaches since they show the best quality. The other results are borrowed from \cite{once}. Detection performance is evaluated using mAp on the validation split. The models trained on the Small, Medium, and Large unlabeled subsets are considered separately. By comparing the performance of these methods over different amounts of unlabeled data, we can gain insight into their effectiveness and scalability in the context of semi-supervised learning for 3D object detection.

The results in Tables \ref{results_tab_second} and \ref{results_tab_center} show that hyperparameter optimization significantly improves model performance with and without applying semi-supervised techniques. Models marked with "*" are trained with our hyperparameters and others with the original ones. In particular, the quality of the pretrained SECOND* model, which uses our hyperparameter setup and hasn't been exposed to any unlabeled data, outperforms the original Proficient Teacher trained with all labeled and unlabeled data. In addition, the gap between supervised pretraining and semi-supervised learning is not as large as expected. Considering the Small subset, it is 3.45 mAP and 5.83 mAP (respectively for Mean Teacher and Proficient) for SECOND and only 1.09 mAP and 3.84 mAP for SECOND*. It shows that some of the performance gains reported in previous works are not the result of utilizing unlabeled data but of prolonging training with labeled data. However, there is still some performance improvement that validates these approaches. Also, a Proficient Teacher, specifically designed for 3D object detection, steadily surpasses a general approach Mean Teacher. It demonstrates that the research aimed at specifying semi-supervised methods for different tasks is fruitful and encourages further exploration.

\section{Conclusion}
In this study, we have addressed the increasing demand for advanced deep learning technologies in 3D data detection, crucial in domains like autonomous driving, robotics, and augmented reality. Focusing on LiDAR-based point cloud representation for its reliability, we have considered reducing the 3D data annotation workload through semi-supervised learning.
We have highlighted the importance of diverse datasets and identified the ONCE dataset as a key benchmark for evaluating semi-supervised 3D object detection models. Our findings demonstrate the critical role of a well-fitted pretrained model in the success of semi-supervised learning.
Our primary focus has been optimizing supervised pretraining. We have refined the SECOND and CenterPoint models through parameter optimization, improving benchmark results, achieving state-of-the-art performance for these models, and documenting metrics for semi-supervised methods.

\bibliographystyle{ieeetr}
\bibliography{main}

\begin{thebibliography}{10}

\bibitem{solid_lidar}
N.~Li, C.~P. Ho, J.~Xue, L.~W. Lim, G.~Chen, Y.~H. Fu, and L.~Y.~T. Lee, ``A progress review on solid-state lidar and nanophotonics-based lidar sensors,'' {\em Laser \& Photonics Reviews}, vol.~16, no.~11, p.~2100511, 2022.

\bibitem{wearable_sensors}
A.~Mahmoud, H.~Sadruddin, P.~Coser, and M.~Atia, ``Integration of wearable sensors measurements for indoor pedestrian tracking,'' {\em IEEE Instrumentation \& Measurement Magazine}, vol.~25, no.~1, pp.~46--54, 2022.

\bibitem{lidar_mapping}
C.~K. Toth, ``R\&d of mobile lidar mapping and future trends,'' in {\em Proc. ASPRS Annu. Conf}, pp.~9--13, 2009.

\bibitem{ssl_survey}
X.~Yang, Z.~Song, I.~King, and Z.~Xu, ``A survey on deep semi-supervised learning,'' {\em IEEE Transactions on Knowledge and Data Engineering}, 2022.

\bibitem{once}
J.~Mao, M.~Niu, C.~Jiang, H.~Liang, J.~Chen, X.~Liang, Y.~Li, C.~Ye, W.~Zhang, Z.~Li, {\em et~al.}, ``One million scenes for autonomous driving: Once dataset,'' {\em arXiv preprint arXiv:2106.11037}, 2021.

\bibitem{ProfTeachers}
J.~Yin, J.~Fang, D.~Zhou, L.~Zhang, C.-Z. Xu, J.~Shen, and W.~Wang, ``Semi-supervised 3d object detection with proficient teachers,'' in {\em European Conference on Computer Vision}, pp.~727--743, Springer, 2022.

\bibitem{second}
Y.~Yan, Y.~Mao, and B.~Li, ``Second: Sparsely embedded convolutional detection,'' {\em Sensors}, vol.~18, no.~10, p.~3337, 2018.

\bibitem{centerpoint}
T.~Yin, X.~Zhou, and P.~Krahenbuhl, ``Center-based 3d object detection and tracking,'' in {\em Proceedings of the IEEE/CVF conference on computer vision and pattern recognition}, pp.~11784--11793, 2021.

\bibitem{MeanTeachers}
H.~V. Antti~Tarvainen, ``Mean teachers are better role models: Weight-averaged consistency targets improve semi-supervised deep learning results.,'' 2017.

\bibitem{rcnn}
R.~Girshick, J.~Donahue, T.~Darrell, and J.~Malik, ``Rich feature hierarchies for accurate object detection and semantic segmentation,'' in {\em Proceedings of the IEEE Conference on Computer Vision and Pattern Recognition (CVPR)}, June 2014.

\bibitem{fast_rcnn}
R.~Girshick, ``Fast r-cnn,'' in {\em Proceedings of the IEEE International Conference on Computer Vision (ICCV)}, December 2015.

\bibitem{faster_rcnn}
S.~Ren, K.~He, R.~Girshick, and J.~Sun, ``Faster r-cnn: Towards real-time object detection with region proposal networks,'' {\em Advances in neural information processing systems}, vol.~28, 2015.

\bibitem{rfcn}
J.~Dai, Y.~Li, K.~He, and J.~Sun, ``R-fcn: Object detection via region-based fully convolutional networks,'' {\em Advances in neural information processing systems}, vol.~29, 2016.

\bibitem{fpn}
T.-Y. Lin, P.~Doll{\'a}r, R.~Girshick, K.~He, B.~Hariharan, and S.~Belongie, ``Feature pyramid networks for object detection,'' in {\em Proceedings of the IEEE conference on computer vision and pattern recognition}, pp.~2117--2125, 2017.

\bibitem{cascade_rcnn}
Z.~Cai and N.~Vasconcelos, ``Cascade r-cnn: Delving into high quality object detection,'' in {\em Proceedings of the IEEE conference on computer vision and pattern recognition}, pp.~6154--6162, 2018.

\bibitem{libra_rcnn}
J.~Pang, K.~Chen, J.~Shi, H.~Feng, W.~Ouyang, and D.~Lin, ``Libra r-cnn: Towards balanced learning for object detection,'' in {\em Proceedings of the IEEE/CVF Conference on Computer Vision and Pattern Recognition (CVPR)}, 2019.

\bibitem{htc}
K.~Chen, J.~Pang, J.~Wang, Y.~Xiong, X.~Li, S.~Sun, W.~Feng, Z.~Liu, J.~Shi, W.~Ouyang, C.~C. Loy, and D.~Lin, ``Hybrid task cascade for instance segmentation,'' in {\em Proceedings of the IEEE/CVF Conference on Computer Vision and Pattern Recognition (CVPR)}, 2019.

\bibitem{detectors}
S.~Qiao, L.-C. Chen, and A.~Yuille, ``Detectors: Detecting objects with recursive feature pyramid and switchable atrous convolution,'' in {\em Proceedings of the IEEE/CVF Conference on Computer Vision and Pattern Recognition (CVPR)}, pp.~10213--10224, 2021.

\bibitem{yolov1}
J.~Redmon, S.~Divvala, R.~Girshick, and A.~Farhadi, ``You only look once: Unified, real-time object detection,'' in {\em Proceedings of the IEEE conference on computer vision and pattern recognition}, pp.~779--788, 2016.

\bibitem{yolov2}
J.~Redmon and A.~Farhadi, ``Yolo9000: better, faster, stronger,'' in {\em Proceedings of the IEEE conference on computer vision and pattern recognition}, pp.~7263--7271, 2017.

\bibitem{yolov3}
J.~Redmon and A.~Farhadi, ``Yolov3: An incremental improvement,'' {\em arXiv}, 2018.

\bibitem{focal_loss}
T.-Y. Lin, P.~Goyal, R.~Girshick, K.~He, and P.~Doll{\'a}r, ``Focal loss for dense object detection,'' in {\em Proceedings of the IEEE international conference on computer vision}, pp.~2980--2988, 2017.

\bibitem{ssd}
W.~Liu, D.~Anguelov, D.~Erhan, C.~Szegedy, S.~Reed, C.-Y. Fu, and A.~C. Berg, ``Ssd: Single shot multibox detector,'' in {\em Computer Vision--ECCV 2016: 14th European Conference, Amsterdam, The Netherlands, October 11--14, 2016, Proceedings, Part I 14}, pp.~21--37, Springer, 2016.

\bibitem{densebox}
L.~Huang, Y.~Yang, Y.~Deng, and Y.~Yu, ``Densebox: Unifying landmark localization with end to end object detection,'' {\em arXiv preprint arXiv:1509.04874}, 2015.

\bibitem{fsaf}
C.~Zhu, Y.~He, and M.~Savvides, ``Feature selective anchor-free module for single-shot object detection,'' in {\em Proceedings of the IEEE/CVF conference on computer vision and pattern recognition}, pp.~840--849, 2019.

\bibitem{detr}
N.~Carion, F.~Massa, G.~Synnaeve, N.~Usunier, A.~Kirillov, and S.~Zagoruyko, ``End-to-end object detection with transformers,'' in {\em European conference on computer vision}, pp.~213--229, Springer, 2020.

\bibitem{reppoints}
Z.~Yang, S.~Liu, H.~Hu, L.~Wang, and S.~Lin, ``Reppoints: Point set representation for object detection,'' in {\em Proceedings of the IEEE/CVF international conference on computer vision}, pp.~9657--9666, 2019.

\bibitem{fcos}
Z.~Tian, C.~Shen, H.~Chen, and T.~He, ``Fcos: Fully convolutional one-stage object detection,'' in {\em Proceedings of the IEEE/CVF international conference on computer vision}, pp.~9627--9636, 2019.

\bibitem{fovea}
T.~Kong, F.~Sun, H.~Liu, Y.~Jiang, L.~Li, and J.~Shi, ``Foveabox: Beyound anchor-based object detection,'' {\em IEEE Transactions on Image Processing}, pp.~7389--7398, 2020.

\bibitem{cornernet}
H.~Law and J.~Deng, ``Cornernet: Detecting objects as paired keypoints,'' in {\em Proceedings of the European conference on computer vision (ECCV)}, pp.~734--750, 2018.

\bibitem{extremenet}
X.~Zhou, J.~Zhuo, and P.~Krahenbuhl, ``Bottom-up object detection by grouping extreme and center points,'' in {\em Proceedings of the IEEE/CVF conference on computer vision and pattern recognition}, pp.~850--859, 2019.

\bibitem{pixor}
B.~Yang, W.~Luo, and R.~Urtasun, ``Pixor: Real-time 3d object detection from point clouds,'' in {\em Proceedings of the IEEE conference on Computer Vision and Pattern Recognition}, pp.~7652--7660, 2018.

\bibitem{pdv}
J.~S. Hu, T.~Kuai, and S.~L. Waslander, ``Point density-aware voxels for lidar 3d object detection,'' in {\em Proceedings of the IEEE/CVF Conference on Computer Vision and Pattern Recognition}, pp.~8469--8478, 2022.

\bibitem{sst}
L.~Fan, Z.~Pang, T.~Zhang, Y.-X. Wang, H.~Zhao, F.~Wang, N.~Wang, and Z.~Zhang, ``Embracing single stride 3d object detector with sparse transformer,'' in {\em Proceedings of the IEEE/CVF conference on computer vision and pattern recognition}, pp.~8458--8468, 2022.

\bibitem{centernet}
K.~Duan, S.~Bai, L.~Xie, H.~Qi, Q.~Huang, and Q.~Tian, ``Centernet: Keypoint triplets for object detection,'' in {\em Proceedings of the IEEE/CVF international conference on computer vision}, pp.~6569--6578, 2019.

\bibitem{centerformer}
Z.~Zhou, X.~Zhao, Y.~Wang, P.~Wang, and H.~Foroosh, ``Centerformer: Center-based transformer for 3d object detection,'' in {\em European Conference on Computer Vision}, pp.~496--513, Springer, 2022.

\bibitem{liga_stereo}
X.~Guo, S.~Shi, X.~Wang, and H.~Li, ``Liga-stereo: Learning lidar geometry aware representations for stereo-based 3d detector,'' in {\em Proceedings of the IEEE/CVF International Conference on Computer Vision}, pp.~3153--3163, 2021.

\bibitem{p2s}
Y.~Xue, J.~Mao, M.~Niu, H.~Xu, M.~B. Mi, W.~Zhang, X.~Wang, and X.~Wang, ``Point2seq: Detecting 3d objects as sequences,'' in {\em Proceedings of the IEEE/CVF Conference on Computer Vision and Pattern Recognition}, pp.~8521--8530, 2022.

\bibitem{vista}
C.~Park, Y.~Jeong, M.~Cho, and J.~Park, ``Fast point transformer,'' in {\em Proceedings of the IEEE/CVF Conference on Computer Vision and Pattern Recognition}, pp.~16949--16958, 2022.

\bibitem{btc}
Q.~Xu, Y.~Zhong, and U.~Neumann, ``Behind the curtain: Learning occluded shapes for 3d object detection,'' in {\em Proceedings of the AAAI Conference on Artificial Intelligence}, vol.~36, pp.~2893--2901, 2022.

\bibitem{voxelnet}
Y.~Zhou and O.~Tuzel, ``Voxelnet: End-to-end learning for point cloud based 3d object detection,'' in {\em Proceedings of the IEEE conference on computer vision and pattern recognition}, pp.~4490--4499, 2018.

\bibitem{pointpillars}
A.~H. Lang, S.~Vora, H.~Caesar, L.~Zhou, J.~Yang, and O.~Beijbom, ``Pointpillars: Fast encoders for object detection from point clouds,'' in {\em Proceedings of the IEEE/CVF conference on computer vision and pattern recognition}, pp.~12697--12705, 2019.

\bibitem{lidar_rcnn}
Z.~Li, F.~Wang, and N.~Wang, ``Lidar r-cnn: An efficient and universal 3d object detector,'' in {\em Proceedings of the IEEE/CVF Conference on Computer Vision and Pattern Recognition}, pp.~7546--7555, 2021.

\bibitem{3dssd}
Z.~Yang, Y.~Sun, S.~Liu, and J.~Jia, ``3dssd: Point-based 3d single stage object detector,'' in {\em Proceedings of the IEEE/CVF conference on computer vision and pattern recognition}, pp.~11040--11048, 2020.

\bibitem{ia_ssd}
Y.~Zhang, Q.~Hu, G.~Xu, Y.~Ma, J.~Wan, and Y.~Guo, ``Not all points are equal: Learning highly efficient point-based detectors for 3d lidar point clouds,'' in {\em Proceedings of the IEEE/CVF Conference on Computer Vision and Pattern Recognition}, pp.~18953--18962, 2022.

\bibitem{pointnet}
C.~R. Qi, H.~Su, K.~Mo, and L.~J. Guibas, ``Pointnet: Deep learning on point sets for 3d classification and segmentation,'' in {\em Proceedings of the IEEE conference on computer vision and pattern recognition}, pp.~652--660, 2017.

\bibitem{pointnetpp}
C.~R. Qi, L.~Yi, H.~Su, and L.~J. Guibas, ``Pointnet++: Deep hierarchical feature learning on point sets in a metric space,'' {\em Advances in neural information processing systems}, vol.~30, 2017.

\bibitem{frustum_pointnet}
C.~R. Qi, W.~Liu, C.~Wu, H.~Su, and L.~J. Guibas, ``Frustum pointnets for 3d object detection from rgb-d data,'' in {\em Proceedings of the IEEE conference on computer vision and pattern recognition}, pp.~918--927, 2018.

\bibitem{point_rcnn}
S.~Shi, X.~Wang, and H.~Li, ``Pointrcnn: 3d object proposal generation and detection from point cloud,'' in {\em Proceedings of the IEEE/CVF conference on computer vision and pattern recognition}, pp.~770--779, 2019.

\bibitem{votenet}
C.~R. Qi, O.~Litany, K.~He, and L.~J. Guibas, ``Deep hough voting for 3d object detection in point clouds,'' in {\em proceedings of the IEEE/CVF International Conference on Computer Vision}, pp.~9277--9286, 2019.

\bibitem{vst}
C.~He, R.~Li, S.~Li, and L.~Zhang, ``Voxel set transformer: A set-to-set approach to 3d object detection from point clouds,'' in {\em Proceedings of the IEEE/CVF Conference on Computer Vision and Pattern Recognition}, pp.~8417--8427, 2022.

\bibitem{fpt}
C.~Park, Y.~Jeong, M.~Cho, and J.~Park, ``Fast point transformer,'' in {\em Proceedings of the IEEE/CVF Conference on Computer Vision and Pattern Recognition}, pp.~16949--16958, 2022.

\bibitem{std}
Z.~Yang, Y.~Sun, S.~Liu, X.~Shen, and J.~Jia, ``Std: Sparse-to-dense 3d object detector for point cloud,'' in {\em Proceedings of the IEEE/CVF international conference on computer vision}, pp.~1951--1960, 2019.

\bibitem{pv_rcnn}
S.~Shi, C.~Guo, L.~Jiang, Z.~Wang, J.~Shi, X.~Wang, and H.~Li, ``Pv-rcnn: Point-voxel feature set abstraction for 3d object detection,'' in {\em Proceedings of the IEEE/CVF conference on computer vision and pattern recognition}, pp.~10529--10538, 2020.

\bibitem{hvpr}
J.~Noh, S.~Lee, and B.~Ham, ``Hvpr: Hybrid voxel-point representation for single-stage 3d object detection,'' in {\em Proceedings of the IEEE/CVF conference on computer vision and pattern recognition}, pp.~14605--14614, 2021.

\bibitem{smoothness}
A.~Oliver, A.~Odena, C.~A. Raffel, E.~D. Cubuk, and I.~Goodfellow, ``Realistic evaluation of deep semi-supervised learning algorithms,'' {\em Advances in neural information processing systems}, vol.~31, 2018.

\bibitem{noisy_student}
Q.~Xie, M.-T. Luong, E.~Hovy, and Q.~V. Le, ``Self-training with noisy student improves imagenet classification,'' in {\em Proceedings of the IEEE/CVF conference on computer vision and pattern recognition}, pp.~10687--10698, 2020.

\bibitem{fixmatch}
K.~Sohn, D.~Berthelot, N.~Carlini, Z.~Zhang, H.~Zhang, C.~A. Raffel, E.~D. Cubuk, A.~Kurakin, and C.-L. Li, ``Fixmatch: Simplifying semi-supervised learning with consistency and confidence,'' {\em Advances in neural information processing systems}, vol.~33, pp.~596--608, 2020.

\bibitem{conf_bias}
E.~Arazo, D.~Ortego, P.~Albert, N.~E. O’Connor, and K.~McGuinness, ``Pseudo-labeling and confirmation bias in deep semi-supervised learning,'' in {\em 2020 International Joint Conference on Neural Networks (IJCNN)}, pp.~1--8, 2020.

\bibitem{simmatch}
M.~Zheng, S.~You, L.~Huang, F.~Wang, C.~Qian, and C.~Xu, ``Simmatch: Semi-supervised learning with similarity matching,'' in {\em Proceedings of the IEEE/CVF Conference on Computer Vision and Pattern Recognition}, pp.~14471--14481, 2022.

\bibitem{cossl}
Y.~Fan, D.~Dai, A.~Kukleva, and B.~Schiele, ``Cossl: Co-learning of representation and classifier for imbalanced semi-supervised learning,'' in {\em Proceedings of the IEEE/CVF conference on computer vision and pattern recognition}, pp.~14574--14584, 2022.

\bibitem{temp_ensembling}
S.~Laine and T.~Aila, ``Temporal ensembling for semi-supervised learning,'' {\em arXiv preprint arXiv:1610.02242}, 2016.

\bibitem{mixmatch}
D.~Berthelot, N.~Carlini, I.~Goodfellow, N.~Papernot, A.~Oliver, and C.~A. Raffel, ``Mixmatch: A holistic approach to semi-supervised learning,'' {\em Advances in neural information processing systems}, vol.~32, 2019.

\bibitem{csd}
J.~Jeong, S.~Lee, J.~Kim, and N.~Kwak, ``Consistency-based semi-supervised learning for object detection,'' {\em Advances in neural information processing systems}, vol.~32, 2019.

\bibitem{isd}
J.~Jeong, V.~Verma, M.~Hyun, J.~Kannala, and N.~Kwak, ``Interpolation-based semi-supervised learning for object detection,'' in {\em Proceedings of the IEEE/CVF Conference on Computer Vision and Pattern Recognition}, pp.~11602--11611, 2021.

\bibitem{stac}
K.~Sohn, Z.~Zhang, C.-L. Li, H.~Zhang, C.-Y. Lee, and T.~Pfister, ``A simple semi-supervised learning framework for object detection,'' {\em arXiv preprint arXiv:2005.04757}, 2020.

\bibitem{ioumatch}
H.~Wang, Y.~Cong, O.~Litany, Y.~Gao, and L.~J. Guibas, ``3dioumatch: Leveraging iou prediction for semi-supervised 3d object detection,'' in {\em Proceedings of the IEEE/CVF Conference on Computer Vision and Pattern Recognition}, pp.~14615--14624, 2021.

\bibitem{lr_bs}
S.~L. Smith, P.-J. Kindermans, C.~Ying, and Q.~V. Le, ``Don't decay the learning rate, increase the batch size,'' {\em arXiv preprint arXiv:1711.00489}, 2017.

\bibitem{pascalvoc}
M.~Everingham, L.~Van~Gool, C.~K. Williams, J.~Winn, and A.~Zisserman, ``The pascal visual object classes (voc) challenge,'' {\em International journal of computer vision}, vol.~88, pp.~303--338, 2010.

\bibitem{one_cycle}
L.~N. Smith and N.~Topin, ``Super-convergence: Very fast training of neural networks using large learning rates,'' in {\em Artificial intelligence and machine learning for multi-domain operations applications}, vol.~11006, pp.~369--386, SPIE, 2019.

\bibitem{pseudolabel}
D.-H. Lee {\em et~al.}, ``Pseudo-label: The simple and efficient semi-supervised learning method for deep neural networks,'' in {\em Workshop on challenges in representation learning, ICML}, vol.~3, p.~896, Atlanta, 2013.

\bibitem{sess}
N.~Zhao, T.-S. Chua, and G.~H. Lee, ``Sess: Self-ensembling semi-supervised 3d object detection,'' in {\em Proceedings of the IEEE/CVF Conference on Computer Vision and Pattern Recognition}, pp.~11079--11087, 2020.

\end{thebibliography}



%% For example:

%% \bibitem[Aladro et al.(2015)]{Aladro15} Aladro, R., Martín, S., Riquelme, D., et al. 2015, \aas, 579, A101

%%\end{thebibliography}

\end{document}